**An Application of a Multivariate Estimation of Distribution Algorithm to Cancer Chemotherapy**

Alexander Brownlee, Martin Pelikan, John McCall, and Andrei Petrovski

# An Application of a Multivariate Estimation of Distribution Algorithm to Cancer Chemotherapy

Alexander Brownlee
School of Computing,
The Robert Gordon University
Aberdeen, UK
sb@comp.rgu.ac.uk

Martin Pelikan
Dept. of Mathematics and Computer Science,
University of Missouri
St. Louis, USA
pelikan@cs.umsl.edu

John McCall
School of Computing,
The Robert Gordon University
Aberdeen, UK
jm@comp.rgu.ac.uk

Andrei Petrovski
School of Computing,
The Robert Gordon University
Aberdeen, UK
ap@comp.rgu.ac.uk

## ABSTRACT
Chemotherapy treatment for cancer is a complex optimisation problem with a large number of interacting variables and constraints. A number of different probabilistic algorithms have been applied to it with varying success. In this paper we expand on this by applying two estimation of distribution algorithms to the problem. One is UMDA, which uses a univariate probabilistic model similar to previously applied EDAs. The other is hBOA, the first EDA using a multivariate probabilistic model to be applied to the chemotherapy problem. While instinct would lead us to predict that the more sophisticated algorithm would yield better performance on a complex problem like this, we show that it is outperformed by the algorithms using the simpler univariate model. We hypothesise that this is caused by the more sophisticated algorithm being impeded by the large number of interactions in the problem which are unnecessary for its solution.

GECCO Track: "Estimation of Distribution Algorithms"

## Categories and Subject Descriptors
I.2.8 [**Artificial Intelligence**]: Problem Solving, Control Methods, and Search; G.3 [**Probability and Statistics**]: Probabilistic Algorithms, stochastic processes; J.3 [**Life and Medical Sciences**]: Health

## General Terms
Algorithms, Performance, Theory.

## Keywords
Estimation of Distribution Algorithms, Complexity, Medical Applications.

## 1. INTRODUCTION
Chemotherapy treatment for cancer is a complex optimisation problem with a large number of variables and constraints. A delicate balance of drug doses must be struck which controls tumour size while minimising toxic side effects. Finding a suitable treatment regime from the large number of possibilities can be interpreted as a classic search problem suitable for evolutionary algorithms.

Previous research [9, 16, 17, 18, 19] has shown that a large range of probabilistic algorithms can be applied to the chemotherapy problem with varying degrees of success – these have included genetic algorithms, particle swarm optimization and estimation of distribution algorithms (EDAs). Related work applying evolutionary algorithms to chemotherapy can also be found in [1, 6, 7, 11, 20, 21, 22].

Estimation of distribution algorithms are a development of genetic algorithms, replacing the traditional crossover and mutation operators with the building and sampling of a probabilistic model. They may be classified as univariate, bivariate or multivariate according to the degree of interaction between variables in this model [12]. Previous applications of EDAs to chemotherapy have used algorithms employing a univariate model [17]. Here we show the application of a multivariate EDA (hBOA) and a further univariate EDA (UMDA) to the problem and report the performance of each. We then compare this to previously reported results. We see that for this problem, despite its complexity, the EDAs using a univariate model outperform that using a multivariate model. This is in addition to the higher overhead of the more complex algorithm. We draw the conclusion that this is because the multivariate algorithm is distracted by interactions between problem variables which are unnecessary for its solution.

The rest of this paper is structured as follows. In section 2 we describe in detail the nature of the chemotherapy problem. In section 3 we present the experimental procedure used and algorithms which were applied. Section 4 contains the results of the experiments, with analysis of these results given in section 5.

## 2. CANCER CHEMOTHERAPY
Amongst the modalities of cancer treatment, chemotherapy is often considered as inherently the most complex [23]. As a consequence of this, it is extremely difficult to find effective chemotherapy treatments without a systematic approach. In order to realise such an approach, we need to take into account the medical aspects of cancer treatment.

### 2.1 Medical aspects of Chemotherapy
Drugs used in cancer chemotherapy all have narrow therapeutic indices. This means that the dose levels at which these drugs significantly affect a tumour are close to those levels at which unacceptable toxic side-effects occur. Therefore, more effective treatments result from balancing the beneficial and adverse effects of a combination of different drugs, administered at various dosages over a treatment period [19]. The beneficial effects of cancer chemotherapy correspond to treatment objectives which oncologists want to achieve by means of administering anti-cancer drugs. A cancer chemotherapy treatment may be either curative or palliative. Curative treatments attempt to eradicate the tumour; palliative treatments, on the other hand, are applied only when a tumour is deemed to be incurable with the objective to maintain a reasonable quality of life for as long as possible. The adverse effects of cancer chemotherapy stem from the systemic nature of this treatment: drugs are delivered via the bloodstream and therefore affect all body tissues. Since most anti-cancer drugs are highly toxic, they inevitably cause damage to sensitive tissues elsewhere in the body. In order to limit this damage, toxicity constraints need to be placed on the amount of drug applied at any time interval, on the cumulative drug dosage over the treatment period, and on the damage caused to various sensitive tissues [23]. In addition to toxicity constraints, the tumour size (i.e. the number of cancerous cells) must be maintained below a lethal level during the whole treatment period for obvious reasons. The goal of cancer chemotherapy therefore is to achieve the beneficial effects of treatment objectives without violating any of the above mentioned constraints.

## 2.2 Problem formulation

In order to solve the optimisation problem of cancer chemotherapy, we need to find a set of treatment schedules, which satisfies toxicity and tumour size constraints yielding at the same time acceptable values of treatment objectives. This set will allow the oncologist to make a decision on which treatment schedule to use, given his/her preferences or certain priorities. In the remainder of this section we will follow the approach used in [17] to define the decision vectors and the search space for the cancer chemotherapy optimisation problem, specify the constraints, and particularise the optimisation objectives. Anti-cancer drugs are usually delivered according to a discrete dosage program in which there are s doses given at times $t_1, t_2, ..., t_s$ [8]. In the case of multi-drug chemotherapy, each dose is a cocktail of d drugs characterised by the concentration levels $C_{ij}, i \in \overline{1,s}, j \in \overline{1,d}$, of anti-cancer drugs in the blood plasma. Optimisation of chemotherapeutic treatment is achieved by modification of these variables. Therefore, the solution space $\Omega$ of the chemotherapy optimisation problem is the set of control vectors $c = (C_{ij})$ representing the drug concentration profiles. However, not all of these profiles will be feasible as chemotherapy treatment must be constrained in a number of ways. Although the constraint sets of chemotherapeutic treatment vary from drug to drug as well as with cancer type, they have the following general form.

1. Maximum instantaneous dose $C_{\max}$ for each drug acting as a single agent:

$$g_1(\mathbf{c}) = C_{\max\ j} - C_{ij} \geq 0 \qquad \forall i = \overline{1,n}, j = \overline{1,d} \quad (1)$$

2. Maximum cumulative $C_{cum}$ dose for drug acting as a single agent:

$$g_2(\mathbf{c}) = C_{\text{cum}\ j} - \sum_{i=1}^{n} C_{ij} \geq 0 \qquad \forall j = \overline{1,d} \quad (2)$$

3. Maximum permissible size of the tumour:

$$g_3(\mathbf{c}) = N_{\max} - N(t_i) \geq 0 \qquad \forall i = \overline{1,n} \quad (3)$$

4. Restriction on the toxic side-effects of multi-drug chemotherapy:

$$g_4(\mathbf{c}) = C_{\text{s-eff}\ k} - \sum_{j=1}^{d} \eta_{kj} C_{ij} \geq 0 \qquad \forall i = \overline{1,n}, k = \overline{1,m} \quad (4)$$

The factors $\eta_{kj}$ in the last constraint represent the risk of damaging the $k^{th}$ organ or tissue (such as heart, bone marrow, lung etc.) by administering the $j^{th}$ drug. Estimates of these factors for the drugs most commonly used in treatment of breast cancer, as well as the values of maximum instantaneous and cumulative doses, can be found in [2, 3]. Regarding the objectives of cancer chemotherapy, we focus our study on the primary objective of cancer treatment - tumour eradication. We define eradication to mean a reduction of the tumour from an initial size of around $10^9$ cells (minimum detectable tumour size) to below $10^3$ cells. In order to simulate the response of a tumour to chemotherapy, a number of mathematical models can be used[8]. The most popular is the Gompertz growth model with a linear cell-loss effect [23]:

$$\frac{dN}{dt} = N(t) \cdot \left[ \lambda \ln\left(\frac{\Theta}{N(t)}\right) - \sum_{j=1}^{d} \kappa_j \sum_{i=1}^{n} C_{ij} \{H(t-t_i) - H(t-t_{i+1})\} \right] \quad (5)$$

where $N(t)$ represents the number of tumour cells at time $t$; $\lambda, \Theta$ are the parameters of tumour growth, $H(t)$ is the Heaviside step function; $\kappa_j$ are the quantities representing the efficacy of anti-cancer drugs, and $C_{ij}$ denote the concentration levels of these drugs. One advantage of the Gompertz model from the computational optimisation point of view is that the equation (5) yields an analytical solution after the substitution $u(t) = \ln(\Theta/N(t))$ [8]. Since $u(t)$ increases when $N(t)$ decreases, the primary optimisation objective of tumour eradication can be formulated as follows [16]:

$$\underset{c}{\text{minimise}}\ F(c) = \sum_{i=1}^{s} N(t_i) \quad (6)$$

subject to the state equation (5) and the constraints (1)-(4).

## 2.3 Application of evolutionary algorithms

It has been reported that integer encoding of GA solutions can improve the algorithm's performance [15], but for the purpose of the present study we have implemented all algorithms using binary representation to make the analysis of comparative results valid.

Multi-drug chemotherapy schedules, represented by decision vectors $c = (C_{ij}), i \in \overline{1,s}, j \in \overline{1,d}$, are encoded as binary strings known as chromosomes. The representation space $I$ (a discretised version of $\Omega$) can then be expressed as a Cartesian product

$$I = A_1^1 \times A_1^2 \times \ldots \times A_1^d \times A_2^1 \times A_2^2 \times \ldots \times A_2^d \times \ldots \times A_s^1 \times A_s^2 \times \ldots \times A_s^d$$

of allele sets $A_i^j$. Each allele set uses a 4-bit representation scheme

$$A_i^j = \left\{ x_1 x_2 x_3 x_4 : x_k \in \{0,1\} \forall k = \overline{1,4} \right\}$$

so that each concentration level $C_{ij}$ takes an integer value in the range of 0 to 15 concentration units [19]. In general, with $s$ treatment intervals and up to $2^p$ concentration levels for $d$ drugs, there are up to $2^{spd}$ individual elements. Henceforth we assume that $s = 10$ and that the number of available drugs in restricted to ten [16]. These drugs are delivered sequentially - one after another - to form a multi-drug dose, which is administered periodically over the treatment period that consists of up to $s$ cycles. The values $s = 10$ and $d = 10$ result in the individual (search) space of power $|I| = 2^{400}$ individuals, referred to as chromosomes.

Thus, a chromosome $x \in I$ can be expressed as

$$x = \{x_1 x_2 x_3 ... x_{4sd} : x_k \in \{0,1\} \forall k \in \overline{1, 4sd}\}$$

and the mapping function $m : I \rightarrow C$ between the individual $I$ and the decision vector $C$ spaces can be defined as

$$C_{ij} = \Delta C_j \sum_{k=1}^{4} 2^{4-1} x_{4d(i1)+4(j-1)+k}, \forall i \in \overline{1, s}, j \in \overline{1, d}$$

where $\Delta C_j$ represents the concentration unit for drug $j$.

This function symbolizes the decoding algorithm to derive a decision vector from a chromosome $x$. Applying the evaluation function $F$ to $c$ yields the value of the fitness function for both algorithms.

$$F(c) = \sum_{p=1}^{n} \sum_{j=1}^{d} \kappa_j \sum_{i=1}^{p} C_{ij} e^{\lambda(t_{i-1} - t_p)} - \sum_{s=1}^{4} P_s d_s \quad (7)$$

where $d_s$ are the distance measures specifying how seriously the constraints (1)-(4) are violated, and $P_s$ are the corresponding penalty coefficients. If all constraints are satisfied (i.e. a treatment regimen is feasible), then the second term in (7) will be zero, significantly increasing the value of the fitness function. Thus, a solution which represents a treatment regime in which the patient survives has a raw fitness value greater than zero.

## 2.4 Problem Complexity

Like many real-world problems, the chemotherapy problem has a large number of interactions between variables. We ran the Linkage Detection Algorithm [4] 30 times on the problem to find the interactions (linkage) present between the 400 variables. Of the 79800 possible bivariate interactions, the mean number of interactions discovered was 50491 with a standard deviation of 109.5. This means that to perfectly learn the fitness function would require a highly complex probabilistic model which would be computationally expensive to construct. It must be noted however that this does not mean that a multivariate model is needed for optimisation of this problem.

All algorithms previously applied to the problem have either not used a probabilistic model (such as genetic algorithms) or used a univariate model which ignores these interactions. The optimum for the problem is not known and it is conceivable that better solutions exist than those found previously but which require a more sophisticated search to find. This leads us to consider the application of a multivariate EDA to the problem.

## 3. PROCEDURE

The multivariate EDA we applied to the problem was the Hierarchical Bayesian Optimisation Algorithm (hBOA) [13]. For comparison we also applied a univarate EDA – the Univariate Marginal Distribution Algorithm (UMDA) [10]. The results for these were then compared to those previously published for other algorithms.

For each experiment, the algorithms were run with a number of different parameter configurations to find the optimal performance. Tournament and truncation selection operators were tested with varying proportions of the population being selected. The bisection technique [13] was used to find optimal population size with each configuration. The set of parameters giving the best results over independent 10 runs were then used for the experiments. We now give a brief description of each algorithm with the optimised parameters for each before presenting the experimental results.

### 3.1 Univariate Marginal Distribution Algorithm

UMDA [10] is one of the earliest developed and best-known EDAs. Each generation a number of highly fit individuals are selected from the population. The marginal probabilities for each variable in this set are calculated and then sampled to generate the next population. It may be regarded as PBIL with a learning rate of zero [5]. As a result of using such a simple probabilistic model, it has little algorithm overhead but may converge on sum-optimal solutions.

In the efficiency experiment, UMDA was found to perform best using tournament selection with a pool of 6 individuals, and bisection found an optimum population size of 112. In the solution quality experiment, truncation selection was used with a pool size of 2 and a population size of 800.

### 3.2 Hierarchical Bayesian Optimisation Algorithm

hBOA [13] uses a Bayesian network to factorise the joint probability distribution of good solutions. Previous results have shown it to perform well on complex problems such as MAXSAT and Ising Spin Glasses [14]. hBOA learns the structure of the Bayesian network as the evolution proceeds. This is the first multivariate EDA to be applied to this chemotherapy problem.

For the efficiency experiment, hBOA was determined to perform best when using truncation selection with a pool of 40 individuals and a population size of 400. For the solution quality experiment, truncation selection with a pool size of 2 was used, with a population size of 6400.

### 3.3 Experiments

We ran two experiments with each algorithm:

The first investigated the efficiency or speed of the algorithms. It is often the case with evolutionary algorithms that the fitness function is the most computationally expensive part – especially the case with complex real-world fitness functions. Thus a common measure of performance is the number of fitness evaluations the algorithm requires to find a solution. For this problem, we measure the number of evaluations required to find a feasible solution (that is, one in which all constraints are satisfied and the patient survives).

The second experiment recorded the quality of solution found by each algorithm. In this case, this is the best fitness value found by the algorithm in a fixed number of evaluations (to allow comparison with our previous results this was set to 200 000).

## 4. RESULTS
### 4.1 Efficiency

First we present the number of fitness evaluations taken by each algorithm to reach a feasible solution. The results are the mean of 30 runs. For comparison we also show previously reported results for a genetic algorithm, PBIL and univariate DEUM. These are given in Table 1 and illustrated graphically in Figure 1, where error bars represent one standard deviation.

**Table 1. Mean and standard deviation of fitness evaluations to find a feasible solution**

| Algorithm | Mean number of fitness evaluations | Std Deviation |
|---|---|---|
| UMDA | 2695.5 | 490.3 |
| hBOA | 7917.6 | 843.0 |
| Genetic Algorithm | 16208.1 | 12045.8 |
| PBIL | 5959.5 | 522.4 |
| Univariate DEUM | 5015.8 | 1426.9 |

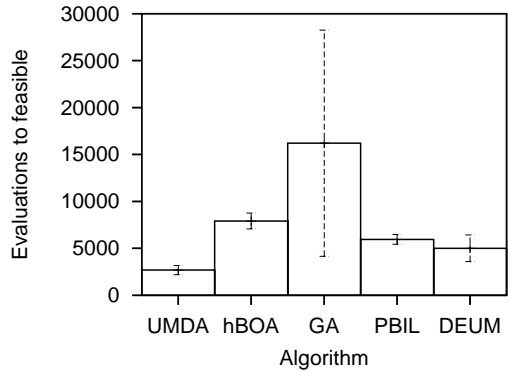

**Figure 1. Graphical comparison of mean fitness evaluations to find a feasible solution**

Table 2 presents the results of T-test analysis on pairs of results given in Table 1. UMDA and hBOA are compared with each other then each is compared in turn with the previously reported results for a GA, PBIL and univariate DEUM. We see that for each pair of results, the p-value is <<0.05 indicating that the difference in means for each pair is statistically significant.

**Table 2. T-test analysis of difference in mean fitness evaluations to find a feasible solution**

|  | Diff in means | Std Error | t-test | p-value |
|---|---|---|---|---|
| UMDA vs hBOA | 5222.1 | 178.049 | 25.2857 | < 0.0001 |
| UMDA vs GA | 13512.6 | 2201.073 | 6.1391 | < 0.0001 |
| UMDA vs PBIL | 3264.0 | 130.805 | 24.9532 | < 0.0001 |
| UMDA vs Univariate DEUM | 2320.4 | 275.466 | 8.4236 | < 0.0001 |
| hBOA vs GA | 8290.5 | 2204.631 | 3.7605 | 0.0004 |
| hBOA vs PBIL | 1958.1 | 181.066 | 10.8143 | < 0.0001 |
| hBOA vs Univariate DEUM | 2901.7 | 302.583 | 9.5898 | < 0.0001 |

## 4.2 Best Fitness Found

In the second experiment we investigate the best fitness found after the number of fitness evaluations has reached 200 000. Again, we also compare with previously reported results for a genetic algorithm, PBIL and univariate DEUM. These results are given in Table 3 and are also illustrated graphically in Figure 2, where error bars represent one standard deviation.

**Table 3. Best fitness found in 200 000 fitness evaluations**

| Algorithm | Mean best fitness found | Std Deviation |
|---|---|---|
| UMDA | 0.4916 | 0.0006 |
| hBOA | 0.4842 | 0.0021 |
| Genetic Algorithm | 0.4520 | 0.0354 |
| PBIL | 0.4917 | 0.0005 |
| Univariate DEUM | 0.4920 | 0.0006 |

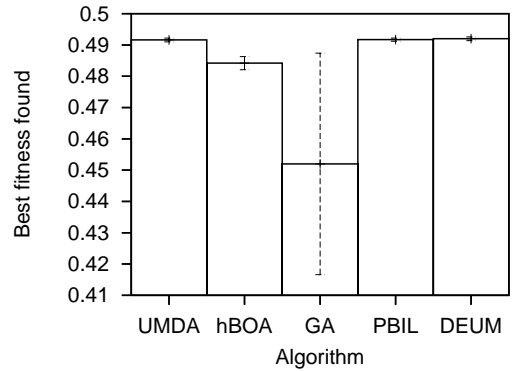

**Figure 2. Graphical comparison of best fitness found in 200 000 fitness evaluations**

In Table 4 we present the results of T-test analysis on pairs of results given in Table 3. As before, UMDA and hBOA are compared with each other then each is compared in turn with the previously reported results for a GA, PBIL and univariate DEUM. For almost all pairs of results the p-value is <<0.05 indicating that the difference in means for each pair is statistically significant. The one exception to this is the difference between PBIL and UMDA – perhaps unsurprising given the similar nature of these algorithms.

**Table 4. T-test analysis of difference in mean fitness evaluations to find a feasible solution**

|  | Diff in means | Std Error | t-test | p-value |
|---|---|---|---|---|
| UMDA vs hBOA | 0.0074 | 0.000 | 18.5581 | < 0.0001 |
| UMDA vs GA | 0.0396 | 0.006 | 6.1262 | < 0.0001 |
| UMDA vs PBIL | 0.0001 | 0.000 | 0.7013 | 0.4859 |
| UMDA vs Univariate DEUM | 0.0004 | 0.000 | 2.5820 | 0.0124 |
| hBOA vs GA | 0.0322 | 0.006 | 4.9734 | < 0.0001 |
| hBOA vs PBIL | 0.0075 | 0.000 | 19.0296 | < 0.0001 |
| hBOA vs Univariate DEUM | 0.0078 | 0.000 | 19.5612 | < 0.0001 |

## 5. ANALYSIS AND CONCLUSIONS

In both experiments we see that the genetic algorithm is outperformed by all of the EDAs. This reinforces the findings of previous comparisons between GA and EDA.

A point worth noting is that the efficiency results given here for UMDA are better than those previously published for PBIL. One possible explanation is that the implementation of UMDA we used also allows tournament selection (not used in the previous work on PBIL). This will likely have an impact on the algorithm's performance.

More interesting is the comparison between the EDAs. As we have seen in section 2.4 the multi-drug chemotherapy problem is highly complex with a large number of interactions between variables. On many other problems with multivariate interactions hBOA has been demonstrated to significantly outperform univariate EDAs [13, 14]. Indeed, in many cases the difference between the algorithms is tractable vs. intractable. Given these facts, it is perhaps unexpected that the univariate algorithms do better than hBOA, both in the number of evaluations and the final fitness reached. We believe that the reason for this is that a large number of the interactions within the problem are unnecessary for optimisation. Thus hBOA constructs a complex model of the problem which is very expensive to search and ultimately counterproductive – in essence being preoccupied by the additional complexity while the simpler algorithms are able to ignore it. The bigger model will also need more computation, both in building the model and in number of solution evaluations. Intuitively we would expect a more precise model of the data to result in a more precise location of optima. However if the benefits gained are small, the computation required to do so may render the algorithm uncompetitive. It would be interesting to further explore the relationship between the costs and benefits of detecting multivariate interactions for other multivariate EDAs and on similar highly multivariate problems.

Further, though a complex problem in itself, this instance of the chemotherapy problem uses a relatively simple model of tumour growth. A more complex model including factors such as more antagonistic drugs and cancers is likely to include higher order interactions. Multivariate EDAs such as hBOA could take advantage of this to give a more pronounced improvement in performance.

It is feasible that given an unlimited number of fitness evaluations the algorithm employing a more complex model will be more capable of finding the true global optimum, but with a huge computational expense. In comparison, the algorithms using a univariate model can get very close to the global optimum – in this case close enough that the reduction in overhead is worthwhile. The addition of a local search operator such a deterministic hillclimber could improve the performance of UMDA and PBIL even further and this would be another area for further investigation.

## 6. ACKNOWLEDGMENT

M.P. was sponsored by the National Science Foundation under CAREER grant ECS-0547013, by the Air Force Office of Scientific Research, Air Force Materiel Command, USAF, under grant FA9550-06-1-0096, and by the University of Missouri in St. Louis through the High Performance Computing Collaboratory sponsored by Information Technology Services, and the Research Award and Research Board programs. The U.S. Government is authorized to reproduce and distribute reprints for government purposes notwithstanding any copyright notation thereon. Any opinions, findings, and conclusions or recommendations expressed in this material are those of the authors and do not necessarily reflect the views of the National Science Foundation, the Air Force Office of Scientific Research, or the U.S. Government. Some experiments were done using the hBOA software developed by Martin Pelikan and David E. Goldberg at the University of Illinois at Urbana-Champaign.